\documentclass[a4]{sab}
\usepackage{fancyhdr} 
\usepackage{graphicx}
\usepackage{epstopdf}
\usepackage{amsmath}
\usepackage{natbib}
\usepackage[pdftex]{hyperref}

\title{Real-time face-swapping as a tool for understanding infant self-recognition}

\affiliation{} 

  \author{Sao Mai Nguyen$^{*}$
        \and
          Masaki Ogino$^{**}$
        \and
          Minoru Asada$^{**}$
}

  \affiliation{
  $^{*}
  $INRIA  Bordeaux - Sud-Ouest, Flowers Team
             \and
  $^{**}
               $Osaka University, Graduate School of Engineering

}

\begin{document}

\maketitle
\thispagestyle{empty}
\thispagestyle{fancy}
\lhead{}
\chead{\vspace{-40pt}
\texttt{\scriptsize{Mai Nguyen, Masaki Ogino, and Minoru Asada. Real-time face-swapping as a tool for understanding infant self-recognition. Proceedings of the 10th International Conference on Epigenetic Robotics, pp.171--172, 2010. }}
\vspace{20pt}}
\rhead{}
\cfoot{}

\begin{abstract}
To study the preference of infants for \textit{contingency of movements} and \textit{familiarity of faces} during self-recognition task, we built, as an accurate and instantaneous imitator,  a real-time face-swapper for videos.

\end{abstract}

\section{Self-recognition development}

Human neonates detect contingency between their movements and what they see \citep{rochat} but they cannot discriminate their own image from that of another infant before 5 months of age. Only by 18 months can they recognize themselves in a mirror. The 6 to18 months period is a decisive developmental stage.

Behavioral studies have shown that 9-month-olds display a preference for familiar faces similar to themselves \citep{sanefuji2}, but also that 5-month-olds show differential visual fixation to a contingent video \citep{bahrick2}.

\section{Unbiased imitator}

\begin{figure}[hbtp]
\centering
  \includegraphics[width = 0.4\textwidth]{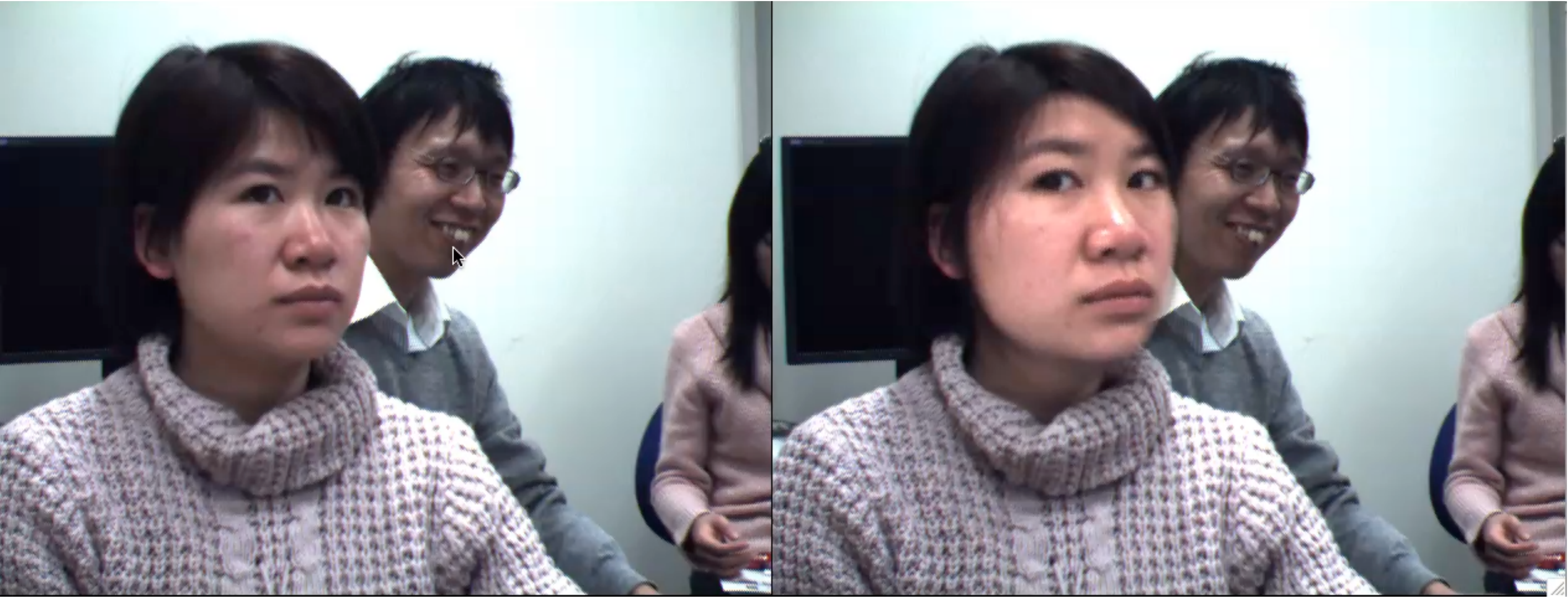}
\caption{face-swap: B's face is superimposed on A's face}
\label{swap}
\end{figure}

We propose to compare the \textit{contingency of movements} and \textit{familiarity of faces} factors in self-recognition in an experiment where an imitator reproduces the head, arms and body movements with or without delay. The imitator's face may be identical to the subject's face or look different.
We thus developed a face-swapper for videos that detects the face position and orientation of the current subject A, then superimposes the image of a subject B on A's face (fig. \ref{swap} where A and B's faces are identical).

To avoid disturbing the subject's behavior or appearance, we did not use special markers. The only installation was a camera. Our \textit{non-constraint real-time} system is an integration of existing 3D head posture trackers, with an original face-swapper in videos. The \textit{very short delay} of the face swapper has been reached thanks to parallel computing including the General-Purpose computing on Graphics Processing Units (GPGPU).  The novelty of this work also lies in its easy calibration. 

\section{Face-swapper for videos}

\begin{figure}[hbtp]
\centering
  \includegraphics[width = 0.5\textwidth]{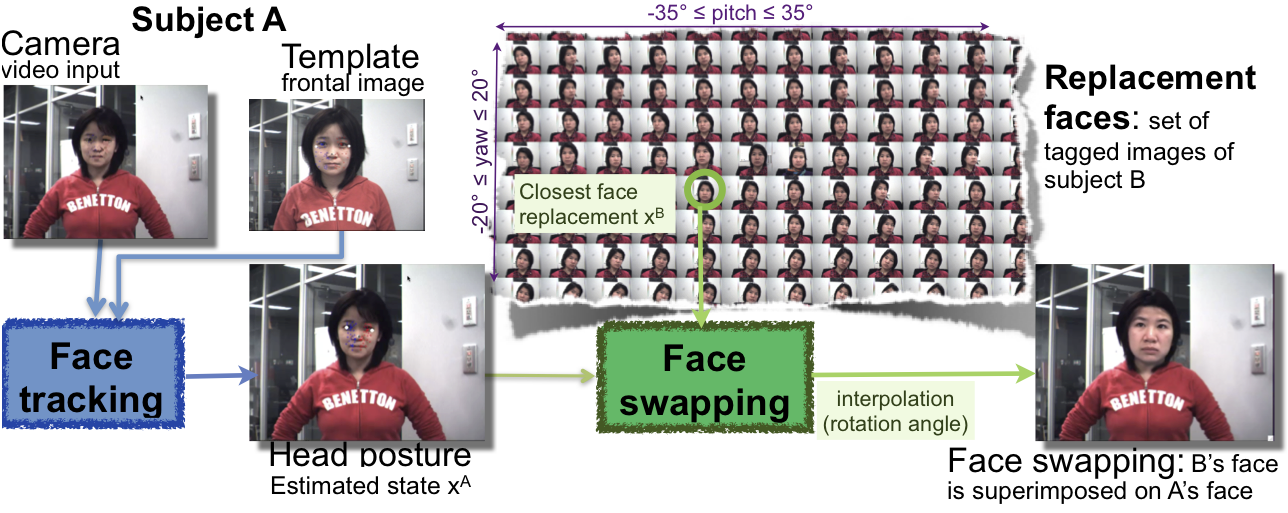}
\caption{The system contains a face tracker and swapper}
\label{overallSystem}
\end{figure}

The overall system (fig. \ref{overallSystem}) includes a \textit{head tracker} to determine the head position and orientation of the current subject A, and a \textit{face swapper} to replace the face of A by that of subject B. Its calibration only uses frontal face pictures of subjects A and B, and the camera video as inputs. 

\subsection{3D visual tracker}
Devices measuring the head's pose such as magnetic sensors, link mechanisms or motion capture unfortunately alter the subjects' behavior  or their natural appearance.
As non-invasive methods, faceAPI, sparse-template-matching-based object tracking \citep{sparse} and CAMSHIFT solutions exist. However they either are commercial systems where the information needed to adapt it for children and extend it to a face-swapper could be inaccessible, or they lack robustness.
 \citet{3DModel} propose an estimation of the 6-DOF motion of the face using a single camera, but require the heavy set-up of a personal 3D facial model.
\citet{visualTracker} present a real-time visual tracker by stream processing and particle filter using a generic 3D model of the face. Our head tracker also adopts this approach to estimate the state
 
 $\textbf{x} =(T_x, T_y,T_{x_{dot}}, T_{y_{dot}}, S, R_x,R_y,R_z,R_{y_{dot}},\alpha)$ 
 
 where 
$T_x, T_y$ are the translation coordinates of the target object,
$T_{x_{dot}}, T_{y_{dot}}$ are the velocity along the horizontal x and vertical y axes,
S is the scale, 
$R_x,R_y,R_z$ are the rotations along each axis,
$R_{y_{dot}}$ is the velocity of the rotation along the vertical axis y, 
and $\alpha$ is a global illumination variable.

Our tracker relies on multi-processing and sparse-template-based particle filtering. No 3D face model was used, but a simpler ellipsoid model. The real-time constraint has been kept thanks to the parallel processing of the camera capture, head-tracking, face-swapping and results-display threads. Moreover the computation of the particle filter was speeded up by the use of GPGPU and NVIDIA CUDA.

\subsection{Face-swapper}
Once $\textbf{x}^{A}$ the face position and orientation of A is detected, an image of B is superimposed on A's face. 

Replacement of whole faces in still images has been developed only recently (\cite{faceAlignment}, \cite{faceSwapping}). 
However, we target videos with real-time constraints, continuity and movement factors.

Our system first creates automatically a set of replacement faces of subject B and tags them with the position and orientation \textbf{x}.
The face-swapper thread compares the state parameters $\textbf{x}^{A}$ with those of the replacement faces of B. It selects the closest face replacement $\textbf{x}^{B}$ to superimpose on A's face. To render the temporal continuity, the replacement face is interpolated  before superimposition, so that the replacement looks dynamic.
We obtain a whole system for automatically replacing faces in videos, that renders dynamic movements of the head. 

\subsection{Performance}
A demonstration video can be found on the site \url{http://www.youtube.com/watch?v=qtYl4o4QoIo}.

 The real-time constraint was the greatest challenge. The use of GPGPU and parallel processing decreased the delay to 99ms.
In addition, our face-swapper is robust against background distractors such as other faces in the background (fig. \ref{swap}). The head-tracker can detect a wide range of face orientations with a head pitch angle up to 70 degrees and is robust against partial occlusion like when children bring their toys or hands to their mouth or faces.
 
The system was evaluated against motion capture system, with an adult subject moving at normal speed, and a head rotation ranging from -40 to 40$^{\circ}$. The pitch angles measured by the motion capture system and ours  show the same variations. The average error is 9$^{\circ}$.

\section{Conclusion}
We presented a non-constraint face-swapper based on 3D visual tracking that achieves real-time performance through parallel computing. 
Our imitator system is particularly suited for experiments involving children with Autistic Spectrum Disorder who are often strongly disturbed by the constraints of other methods \citep{frith}. It can estimate their attention point during natural social interaction to study their peculiar attention pattern, or be used as an imitator to evaluate how imitation facilitates their social behaviors. Future improvement can focus on the facial expressions.

We plan to conduct our experiment with children to investigate the importance of the contingency and familiarity factors in self-recognition. In the longer perspective, the results could be confronted to neuro-scientific data such as the activation in the frontal lobe of the right hemisphere \citep{Uddin} or the default network \citep{Goa} to model the development of self-consciousness, in the context of human self-consciousness analysis, but also for implementation of a robotic sense of self-consciousness.

\textbf{Acknowledgement}: We would like to thank Pierre-Yves Oudeyer for his helpful comments on the draft of this paper and Sawa Fuke for her support for the demonstration video.

{\footnotesize 
\bibliographystyle{sab}
\bibliography{sab_bib_example}
}
\end{document}